\documentclass{article}

\usepackage{microtype}
\usepackage{graphicx}
\usepackage{subcaption} 
\usepackage{booktabs} 
\usepackage{amsmath}
\usepackage{amssymb}
\usepackage{hyperref}

 \usepackage[accepted]{mlsys2025}
 
\mlsystitlerunning{Uncertainty-Aware Multimodal Emotion Recognition through Dirichlet Parameterization}
\begin{document}

\twocolumn[
\mlsystitle{Uncertainty-Aware Multimodal Emotion Recognition through Dirichlet Parameterization}




\begin{mlsysauthorlist}
\mlsysauthor{Rémi Grzeczkowicz}{kaliber}
\mlsysauthor{Eric Soriano}{kaliber}
\mlsysauthor{Ali Janati}{kaliber}
\mlsysauthor{Miyu Zhang}{kaliber}
\mlsysauthor{Gerard Comas-Quiles}{kaliber}
\mlsysauthor{Victor Carballo Araruna}{kaliber}
\mlsysauthor{Aneesh Jonelagadda}{kaliber}
\end{mlsysauthorlist}

\begin{center}
    Kaliber Labs, San Mateo, California, USA
\end{center}
\mlsysaffiliation{kaliber}{Kaliber Labs, San Mateo, California, USA}

\mlsyscorrespondingauthor{Aneesh Jonelagadda}{a.jonelagadda@kaliber.ai}

\mlsyskeywords{Multi Modal Emotion Recognition, Arbitration, Dempster–Shafer}

\vskip 0.3in

\begin{abstract}
In this work, we present a lightweight and privacy-preserving Multimodal Emotion Recognition (MER) framework designed for deployment on edge devices. To demonstrate framework's versatility, our implementation uses three modalities - speech, text and facial imagery. However, the system is fully modular, and can be extended to support other modalities or tasks. Each modality is processed through a dedicated backbone optimized for inference efficiency: Emotion2Vec for speech, a ResNet-based model for facial expressions, and DistilRoBERTa for text.
To reconcile uncertainty across modalities, we introduce a model- and task-agnostic fusion mechanism grounded in Dempster–Shafer theory and Dirichlet evidence. Operating directly on model logits, this approach captures predictive uncertainty without requiring additional training or joint distribution estimation, making it broadly applicable beyond emotion recognition.
Validation on five benchmark datasets (eNTERFACE05, MEAD, MELD, RAVDESS and CREMA-D) show that our method achieves competitive accuracy while remaining computationally efficient and robust to ambiguous or missing inputs. Overall, the proposed framework emphasizes modularity, scalability, and real-world feasibility, paving the way toward uncertainty-aware multimodal systems for healthcare, human–computer interaction, and other emotion-informed applications.
\end{abstract}
]



\printAffiliationsAndNotice{}  

\section{Introduction}
\quad Multimodal Emotion Recognition (MER) seeks to identify human emotions by integrating information from multiple modalities, such as speech, facial expressions, text, and physiological signals. This multimodal fusion enables a richer and more reliable understanding of affective states compared to unimodal approaches that rely on a single data source \cite{abdullah2021multimodal}. By jointly modeling heterogeneous cues—including facial movements, vocal tone, gestures, and linguistic patterns—MER provides a unified framework for capturing and interpreting complex 
emotional dynamics \cite{alswaidan2020survey,xie2013multimodal}.

\quad MER has become increasingly relevant in domains such as healthcare and human–robot interaction, where accurate emotion interpretation enhances user experience and service quality \cite{adel2023mm}. It also underpins the development of empathetic and adaptive AI systems—such as social robots and virtual assistants—capable of responding appropriately to human affect \cite{devault2014simsensei}. MER ultimately attempts to capture a portion of AI system user's behavior. Drawing on Social Psychology frameworks, Behavior ($B$) can be modeled as a function of both the user personality ($P$) and the environment ($E$) of interaction $B=f(P,E)$ \cite{Lewin1936}. Environment is trivial to define, thus maximally understanding user behavior through techniques such as MER allows for distillation of user personality and inherent traits. These traits can be used for downstream tasks such as improving efficacy of future interaction, monitoring, and customization of care.  

\quad The growing range of applications for emotion recognition (ER) highlights the need for robust methodologies that can process diverse emotional cues spanning facial, vocal, physiological, and textual signals \cite{zhao2021emotion}. Recent advances in MER have been driven by progress in machine learning, cross-disciplinary research efforts, and an increased appreciation of the role emotions play in human–computer interaction. Reflecting this momentum, the global ER market was valued at USD~19.87~million in 2020 and is projected to reach USD~52.86~million by 2026, corresponding to a compound annual growth rate (CAGR) of 18.01\% during 2021–2026 \cite{pal2021development}.

For this study, we focus on voice-, text-, and image-based emotion assessments, as these modalities are among the most easily accessible. Since our work targets healthcare applications, patient privacy is a central concern, and our system is designed accordingly. We constrain our approach to computational resources realistically available on edge devices—such as personal computers or home assistants—thereby avoiding reliance on large GPU clusters or cloud-based processing that would require transferring sensitive data externally. Furthermore, our goal extends beyond benchmark performance: rather than following the conventional paradigm of testing on a few datasets with fixed splits, we aim to develop a versatile and edge-ready emotion assessment pipeline capable of operating reliably across diverse, real-world scenarios.

\section{Background and Related Work}

\quad Assessing emotions has been a fundamental research area spanning multiple decades, with applications ranging from human--computer interaction and healthcare to social robotics and market research \cite{abdullah2021multimodal,adel2023mm}. The evolution from unimodal to multimodal emotion recognition represents a significant paradigm shift in affective computing, driven by the understanding that human emotional expressions are inherently multimodal and context-dependent \cite{alswaidan2020survey,xie2013multimodal}.

\subsection{Evolution of Emotion Recognition}

\quad Early emotion recognition research focused primarily on single modalities, such as speech emotion recognition, text emotion recognition, and facial expression recognition \cite{adel2023mm}. However, the limited accuracy of unimodal approaches due to incomplete information and susceptibility to noise led researchers to explore multimodal frameworks \cite{abdullah2021multimodal}. The recognition that humans naturally express emotions through multiple channels simultaneously---facial expressions, vocal cues, linguistic content, and physiological responses---motivated the development of more comprehensive approaches \cite{zhao2021emotion}.

The Circumplex Model of Affect, originally proposed by James Russell in 1980 \cite{russell1980circumplex}, has become one of the most influential theoretical frameworks in emotion recognition. This model represents emotions as combinations of two orthogonal dimensions: valence (pleasantness--unpleasantness) and arousal (activation--deactivation). The circumplex model's two-dimensional representation has proven particularly valuable for computational approaches, as it provides a continuous space for emotion representation rather than discrete categorical labels. It also provides the main 6 emotions: surprise, fear, anger, disgust, sadness, happiness.

\subsection{Speech Emotion Recognition}

\quad Speech emotion recognition has witnessed major advancements through deep learning architectures. Traditional approaches relied on handcrafted acoustic features such as Mel Frequency Cepstral Coefficients (MFCCs) \cite{mishra2024speech}, pitch \cite{koolagudi2012emotion}, and spectral features \cite{koolagudi2009spectral}. Modern approaches leverage sophisticated architectures including Convolutional Neural Networks (CNNs) \cite{huang2014speech}, Long Short-Term Memory (LSTM) networks \cite{zhang2019spontaneous}, and attention mechanisms \cite{lieskovska2021review,tarantino2019self}.

A notable breakthrough came with the introduction of Emotion2Vec, a universal speech emotion representation model that employs self-supervised pre-training on large-scale unlabeled emotion data \cite{ma2023emotion2vec}. Emotion2Vec combines utterance-level and frame-level losses during pre-training and has demonstrated superior performance across multiple languages and emotion-related tasks, achieving state-of-the-art results on the IEMOCAP dataset while requiring only linear layer training for downstream tasks.

\subsection{Facial Emotion Recognition}

\quad Facial emotion recognition has evolved significantly with the development of specialized deep learning architectures optimized for computational efficiency and accuracy trade-offs. Among contemporary approaches, ResEmoteNet represents a notable advancement, achieving 94.76\% accuracy on the RAF-DB dataset with 80.24M parameters \cite{roy2024resemotenet}. The model leverages ResNet-based convolutional architectures with specialized loss reduction techniques to bridge the gap between accuracy and computational overhead.

\quad Large-scale vision transformer approaches have also shown exceptional performance, with FMAE-IAT (Facial Masked Autoencoder with Identity Adversarial Training) achieving 93.54\% accuracy on RAF-DB using a ViT-Large architecture \cite{ning2024representation}. However, such models require substantial computational resources with 304.5M parameters, making them less suitable for real-time applications on edge devices. In contrast, lightweight approaches prioritize efficiency without significant accuracy degradation, as demonstrated by EfficientFace, which achieves 88.28\% accuracy on RAF-DB with only 1.28M parameters \cite{zhao2021robust}.

\quad Attention-based mechanisms have proven particularly effective, with models like DDAMFN (Dual-Direction Attention Mechanism Fusion Network) incorporating CNN architectures with dual-direction attention to achieve 91.35\% accuracy while maintaining moderate computational requirements of 8.2M parameters \cite{zhang2023dual}. Hybrid architectures combining face detection and emotion recognition, such as EasyFace-Emotion, utilize MobileNet for face detection followed by ResNet with attention mechanisms for emotion classification, achieving 89.70\% accuracy on RAF-DB while optimizing the end-to-end pipeline for practical deployment scenarios \cite{githubGitHubSithu31296EasyFace}.

\subsection{Text Emotion Recognition}
\quad Text emotion recognition has been significantly advanced by transformer-based language models, particularly pre-trained language models such as BERT and its variants. Among these, RoBERTa (Robustly Optimized BERT Pretraining Approach) has demonstrated consistently strong performance for emotion classification tasks \cite{liu2019roberta}. Comparative evaluations have shown that RoBERTa and its distilled variants (e.g., DistilRoBERTa) often outperform alternative transformer architectures, including BERT, DistilBERT, and XLNet, in recognizing affective states from text \cite{adoma2020comparative, cortiz2021exploring}. Notably, RoBERTa-based emotion classifiers also dominate widely adopted open-source implementations, reflecting their robustness and practical effectiveness in real-world applications.

\quad The strong performance of transformer models in emotion recognition arises from their ability to model contextual dependencies and capture fine-grained semantic and affective nuances through self-attention mechanisms \cite{barcena2025textual}. Fine-tuning pre-trained language models on emotion-specific datasets—particularly those with fine-grained taxonomies like GoEmotions—has proven highly effective, leveraging the linguistic and contextual representations learned during large-scale pre-training \cite{alqarni2025emotion}.



\subsection{Multimodal Fusion Strategies}

\quad Multimodal emotion recognition systems employ various fusion strategies to combine information from different modalities \cite{karani2022review}. The three primary approaches are:
\begin{itemize}
    \item \textbf{Early Fusion (Feature-level):} Features from different modalities are concatenated before classification \cite{abdullah2021multimodal,alqurashi2024decision}. This approach enables early correlation learning between modalities but may struggle with temporal synchronization issues and modality-specific characteristics.
    \item \textbf{Late Fusion (Decision-level):} Individual modality-specific classifiers make independent predictions, which are then combined using ensemble techniques \cite{alqurashi2024decision,khan2025memocmt}. This approach preserves modality-specific processing but may miss inter-modal correlations.
    \item \textbf{Hybrid Fusion:} Combines elements of both early and late fusion, often achieving superior performance by leveraging the strengths of both approaches.
\end{itemize}

\subsection{Advanced Fusion Techniques}

\quad Recent advances have introduced more sophisticated fusion approaches, including the use of Dempster--Shafer theory for handling uncertainty in multimodal predictions \cite{khan2025memocmt,xu2020emotion}. This framework provides a principled method for combining evidence from multiple modalities while explicitly accounting for uncertainty and potential conflicts between sources.  

\quad Another promising direction is the use of copula-based methods, such as the Student-t copula \cite{ozdemir2017copula}, which can capture complex dependency structures between modalities beyond linear correlations. However, such approaches typically require estimating the correlation matrix of the joint modality distributions to determine the copula parameters. Since our objective is to design a fusion mechanism that generalizes well across datasets and application domains, we deliberately avoid methods that depend on training or data fitting. Instead, we focus on approaches that either operate without additional parameter estimation or remain entirely data-agnostic.

\subsection{Fusionless Strategy}

\quad One primary reason to fuse the modalities is the subsequent ease of mapping to interaction-modulation actions. If the MER dimensionality is low, the mapping mechanism to user-facing reactions such as visual interface changes, audio cues, and delivery can be simply defined. However, more advanced user-facing reactions might be necessary. Thus, as opposed to fusion strategies, another strategy is to persist the different modality outputs into a 3-channel (face, voice, text) descriptor without performing fusion. This higher-dimensional descriptor can still be mapped to user-facing reactions, albeit with higher complexity. Eventually if the dimensionality of the descriptor is high enough, the mapping can be a model to handle concurrent ingestion of additional user-behavior-capturing modalities. In this paper, we choose to implement a fusion approach due to aforementioned simplicity, however future work will involve increasing the dimensionality of behavior descriptors.

\subsection{Benchmark Datasets}
\label{subsec: datasets}

\quad Several benchmark datasets have become standard for evaluating multimodal emotion recognition systems:
\begin{itemize}
    \item \textbf{eNTERFACE05} \cite{martin2006enterface}: An audio-visual emotion database with 42 subjects of 14 nationalities. Participants listened to emotion-eliciting stories and reacted in English. Video was recorded with mini-DV cameras and speech with high-quality microphones. Human annotators selected unambiguous responses. The dataset provides synchronized audio-visual recordings but is relatively small and acted under lab conditions.
    \item \textbf{MEAD (Multi-view Emotional Audio-visual Dataset)} \cite{wang2020mead}: Contains 60 actors recorded simultaneously from seven camera angles. Each actor produced nine emotions (eight basic plus neutral) at three intensity levels (except neutral). The dataset provides high-quality synchronized audio and video with multi-view variation, useful for emotional talking face generation and recognition tasks.
    \item \textbf{MELD (Multimodal EmotionLines Dataset)} \cite{poria2018meld}: Built from \textit{Friends} TV show dialogues, it contains $\sim$1,433 dialogues and over 13k utterances annotated with seven emotions (anger, disgust, sadness, joy, neutral, surprise, fear) and sentiment (positive, negative, neutral). MELD includes text transcripts, audio, and video, making it suitable for multi-party conversational emotion recognition. However, class imbalance (neutral dominates) is a challenge.
    \item \textbf{RAVDESS (Ryerson Audio-Visual Database of Emotional Speech and Song)} \cite{livingstone2018ryerson}: A validated multimodal corpus with 24 actors (12 male, 12 female) producing North American English sentences and songs in eight emotions (neutral, calm, happy, sad, angry, fearful, surprise, disgust). Each non-neutral emotion has two intensity levels (normal, strong). It contains 7,356 files across audio-only, video-only, and audio-visual formats. Emotions were perceptually validated by human raters.
    \item \textbf{CREMA-D (Crowd-sourced Emotional Multimodal Actors Dataset)} \cite{cao2014crema}:  Includes 91 actors (aged 20–74, balanced gender) producing sentences under six emotions (anger, disgust, fear, happy, neutral, sad). It comprises $\sim$7,442 clips of audio-visual recordings, with multiple crowd-sourced annotations of perceived emotion. Its diversity of actors and validation process make it valuable, though emotions remain acted and sentence variety limited.
\end{itemize}





\section{Methodology}
\subsection{Speech Emotion Recognition}

\quad For speech emotion recognition, we employ the Emotion2Vec model \cite{ma2023emotion2vec}. The model is used off-the-shelf and has been trained on a diverse set of datasets covering multiple languages and accents, which enables strong generalization capabilities. Furthermore, the model is relatively lightweight, with approximately 19 million parameters, making it well-suited for deployment on edge devices.

\subsection{Text Emotion Recognition}

\quad For text-based emotion recognition, we employ a DistilRoBERTa model \cite{liu2019roberta,sanh2019distilbert} fine-tuned on a diverse mixture of affective datasets such as MELD \cite{poria2018meld} and GoEmotions \cite{demszky2020goemotions}. The model directly predicts affective states from transcribed utterances without auxiliary stages or intent classification. It has been trained to recognize the basic emotion set (\text{anger}, \text{disgust}, \text{fear}, \text{joy}, \text{neutral}, \text{sadness}, \text{surprise}), following the taxonomy introduced by \cite{ekman1992argument}.

\quad Given an utterance, obtained via the transcription of the audio, the model outputs the raw logits corresponding to each emotion class. This setup preserves the original model architecture while bypassing the softmax normalization step, allowing downstream modules to operate directly on the unscaled logits. Inference is performed using FP16 precision to reduce memory consumption and latency, thereby improving efficiency on low-power devices~\cite{micikevicius2018mixed}.

\quad The model contains approximately 82M parameters and requires only a single forward pass per utterance, making it suitable for real-time, on-device applications.

\subsection{Image Emotion Recognition}
\subsubsection{Image Assessment}
\label{subsubsec:image_assessment}

\quad For image-based emotion recognition, we adopt a two-stage pipeline. In the first stage, facial regions are localized using RetinaFace \cite{deng2020retinaface}, a state-of-the-art face detection framework with approximately 19.72M parameters. The detected faces are then processed by the ResEmotNet model \cite{roy2024resemotenet}, which integrates a ResNet backbone trained on the RAF-DB dataset \cite{li2019reliable}. The model comprises about 80.24M parameters and enables robust feature extraction and emotion classification while maintaining a compact and efficient architecture suitable for deployment on consumer-grade devices.

\subsubsection{Video to Image}

\quad One limitation of the selected approach is that it operates on static images, whereas our setup involves a live-streaming camera (though the video stream can also be divided into individual frames). This raises the problem of selecting a representative frame from the video for emotion assessment.

\quad To address this, we used an additional lightweight image-to-emotion model that evaluates every $T$-th frame and selects the one with the highest saliency, defined as the frame yielding the largest predicted logit magnitude across emotion classes. In our use case, $T$ is set to $5$. Specifically, we employed a ResNet50 architecture \cite{he2016deep} pre-trained on ImageNet-1k \cite{imagenet15russakovsky}, replacing its final layer with a classifier tailored to our emotion categories. The classifier was fine-tuned for two epochs on the RAF-DB dataset \cite{li2019reliable} using the Adam optimizer with a learning rate of $10^{-4}$.

\quad During inference, we retain the frame that yields the highest probability across all emotion classes, defined as:
\[
\text{best\_frame} = \arg\max_{\text{frame}} \big[ \max \big( \text{ResNet50}(\text{frame}) \big) \big].
\]
This selected frame is then passed to the main emotion assessment pipeline described in Section~\ref{subsubsec:image_assessment}. Owing to its reduced size of 23.5M parameters—approximately one-fifth of the main model—this image selector can efficiently operate on edge devices without compromising responsiveness.

\subsection{Arbitration}

\quad Building on Wang et al. ~\yrcite{wang2025uefn}, we consider three modalities: audio, video, and text, denoted by \( n \in \{a, v, t\} \). The task involves classifying inputs into \( K \) emotion categories. We adopt this statistical framework instead of training an additional model, as it is lighter and faster to compute, thereby freeing computing resources for other tasks.

Each modality-specific model outputs logits over the \( K \) classes. We work directly with logits---the raw scores before applying the softmax---since softmax normalizes values into a probability simplex, discarding scale information that is valuable for modeling predictive uncertainty. We denote the logits for modality \( n \) as \( l^n \in \mathbb{R}^K \).

To obtain non-negative \emph{evidence} values, denoted $e$, from the logits, we shift them by subtracting the minimum logit across all modalities, ensuring that the resulting evidence is always zero or positive.
\begin{equation}
e^n = l^n - \min_{m \in \{a,v,t\}} \left( \min l^m \right).
\end{equation}

The evidence is then related to the parameters of the Dirichlet distribution. For each modality,
\begin{equation}
\alpha^n = e^n + 1,
\end{equation}
where both \( e^n \) and \( \alpha^n \) are vectors of dimension \( K \).

The Dirichlet strength \( S^n \) is given by:
\begin{equation}
S^n = \sum_{i=1}^K (e^n_i + 1) = \sum_{i=1}^K \alpha^n_i.
\label{eq:dirichlet-strength}
\end{equation}
\noindent
Intuitively, \( S^n \) represents the \textit{overall confidence} or \textit{total evidence} accumulated by the model for a given sample. A higher \( S^n \) indicates stronger evidence supporting specific classes, while a lower value reflects uncertainty or lack of evidence.

\medskip

We define the belief \( b^n \) as:
\begin{equation}
    b_k^n = \frac{e_k^n}{S^n} = \frac{\alpha_k^n - 1}{S^n}.
\label{eq:belief}
\end{equation}
\noindent
Here, \( b_k^n \) corresponds to the \textit{degree of belief} assigned to class \( k \) after normalizing the available evidence. It quantifies how much support the model attributes to each class relative to the total evidence.

\medskip

The uncertainty \( u^n \) is defined as:
\begin{equation}
    u^n = \frac{K}{S^n}.
\label{eq:uncertainty}
\end{equation}
\noindent
Finally, \( u^n \) captures the model’s \textit{overall uncertainty}, inversely proportional to the Dirichlet strength. When evidence is scarce (\( S^n \) is low), uncertainty increases, reflecting the model’s hesitation in committing to a specific prediction.

Thus, for each modality \( n \), we obtain a result \( M^n \), containing both the belief scores and uncertainty:
\[
M^n = \left\{ \{ b_k^n \}_{k=1}^K,\, u^n \right\}.
\]

To combine results across modalities, we apply Dempster–Shafer theory \cite{Dempster_Shafer_theory}. For two modalities \( M^1 \) and \( M^2 \), the fusion operator \( \oplus \) is defined as:
\begin{equation}
    b_k^1 \oplus b_k^2 = \frac{1}{1 - c} \left( b_k^1 b_k^2 + b_k^1 u^2 + b_k^2 u^1 \right),
\end{equation}
\begin{equation}
    u^1 \oplus u^2 = \frac{1}{1 - c} \left( u^1 u^2 \right),
\end{equation}
where the conflict factor \( c \) is:
\begin{equation}
    c = \sum_{i \neq j} b_i^1 b_j^2.
\end{equation}

The fusion of all three modalities is performed sequentially: we first merge audio and video, \( M^a \oplus M^v \), and then combine the result with text, yielding:
\[
M^{\text{fusion}} = \left\{ \{ b_k \}_{k=1}^K,\, u \right\}.
\]

Finally, we compute the Dirichlet parameters for the fused result. First, the Dirichlet strength is expressed in terms of the uncertainty:
\begin{equation}
    S = \frac{K}{u}.
\end{equation}

The evidence is obtained as:
\begin{equation}
    e_k = b_k  S
\end{equation}

and the Dirichlet parameters follow as:
\begin{equation}
    \alpha_k = e_k + 1.
\end{equation}

The final class probabilities are then given by:
\begin{equation}
    \hat{p}_k = \frac{\alpha_k}{S}.
\label{eq:dirichlet-probabilities}
\end{equation}

\quad This process yields a normalized probability vector \( \hat{p} = [\hat{p}_1, \dots, \hat{p}_K] \) that represents the fused emotional state across modalities. Unlike conventional averaging schemes, this formulation naturally incorporates uncertainty, ensuring that less reliable modalities contribute proportionally less to the final decision. Moreover, this approach is naturally flexible: if a channel is missing, assigning it zero evidence effectively yields an identity operation, allowing the fusion to proceed seamlessly without the need for additional adjustments.

\section{Results}

\begin{figure*}[h!]
    \centering
    \includegraphics[width=0.95\linewidth]{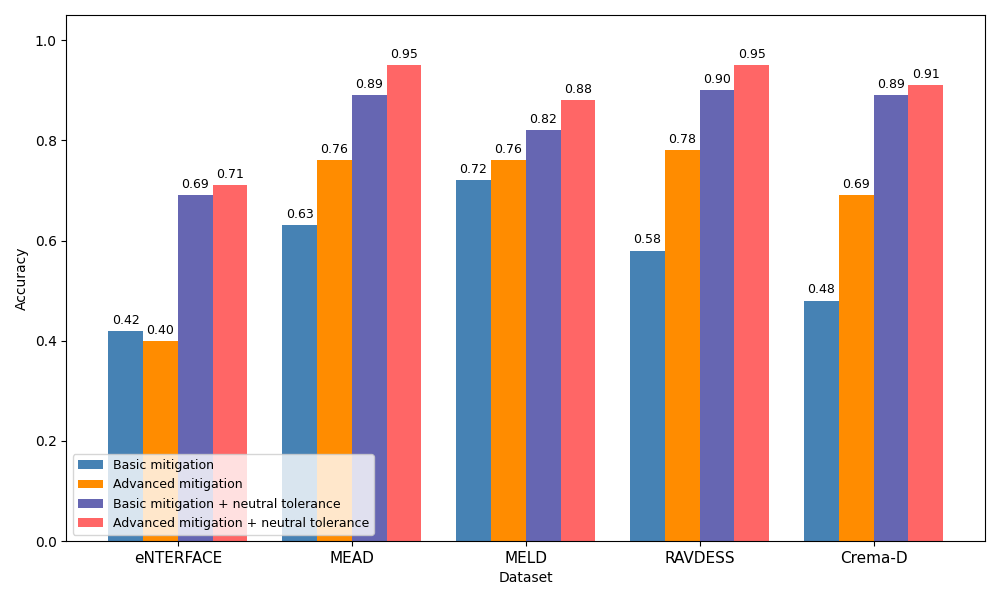}
    \caption{Accuracy with and without neutral tolerance across datasets, using basic and advanced mitigation.}
    \label{fig:accuracy}
\end{figure*}

\quad We tested our pipeline on the five datasets described in Section~\ref{subsec: datasets}. We retained two evaluation metrics: (i) the standard accuracy, and (ii) an adjusted accuracy where predictions of \emph{neutral} are also accepted as correct. This choice was motivated by our intended application: emotions are leveraged to guide interactions with a user, and in this context \emph{neutral} serves as a safe fallback that avoids undesirable misbehavior. Importantly, this adjusted metric is purely for evaluation purposes and does not influence the model’s training, preventing it from collapsing to always predicting \emph{neutral}. Maintaining the standard accuracy alongside the adjusted metric ensures that the model is still incentivized to make specific emotion predictions rather than defaulting to \emph{neutral}.
We compared our proposed mitigation approach against the method of Wang et al. ~\yrcite{wang2025uefn}, which we refer to as \emph{basic mitigation}, while our method is denoted as \emph{advanced mitigation}.

\quad The results summarized in Figure~\ref{fig:accuracy} provides several insights. First, across all datasets our approach consistently outperforms the simple mitigation baseline, with the exception of the eNTERFACE dataset. However, when applying the neutral-tolerant accuracy metric, our method outperforms the baseline on all datasets. Moreover, with neutral tolerance, accuracies reach at least 71\% and are typically closer to 90\%.

\begin{figure}[H]
    \centering
    \includegraphics[width=1\linewidth]{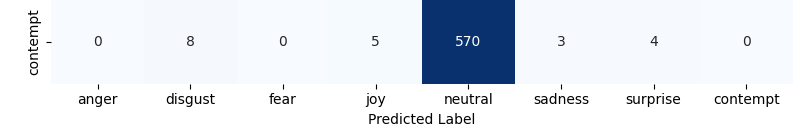}
    \caption{Prediction of \emph{contempt}, an unseen emotion, using our advanced mitigation.}
    \label{fig:mead_contempt}
\end{figure}

\quad More specifically, our approach achieves several important objectives. A key outcome is that \emph{neutral} can now serve as a baseline fallback rather than being treated as a misclassification. For instance, in the MEAD dataset, the \emph{contempt} label—which is absent from the training data and not part of the prediction space—was  mapped to \emph{neutral} $97.9\%$ of the time (see Figure \ref{fig:mead_contempt}). This behavior is desirable, as \emph{neutral} was explicitly designed to act as the fallback emotion. In the RAVDESS and CREMA-D datasets, where the textual content is neutral and emotions are conveyed exclusively through facial expressions or vocal tone, our advanced mitigation substantially improved performance, further underscoring its suitability for multimodal emotion recognition. This advantage is particularly evident in challenging scenarios such as irony, where the literal meaning of the text may contradict the intended emotion; in such cases, our approach demonstrates greater robustness.

\begin{figure}[h]
    \centering
    \begin{subfigure}[b]{0.49\textwidth}
        \centering
        \includegraphics[width=\linewidth]{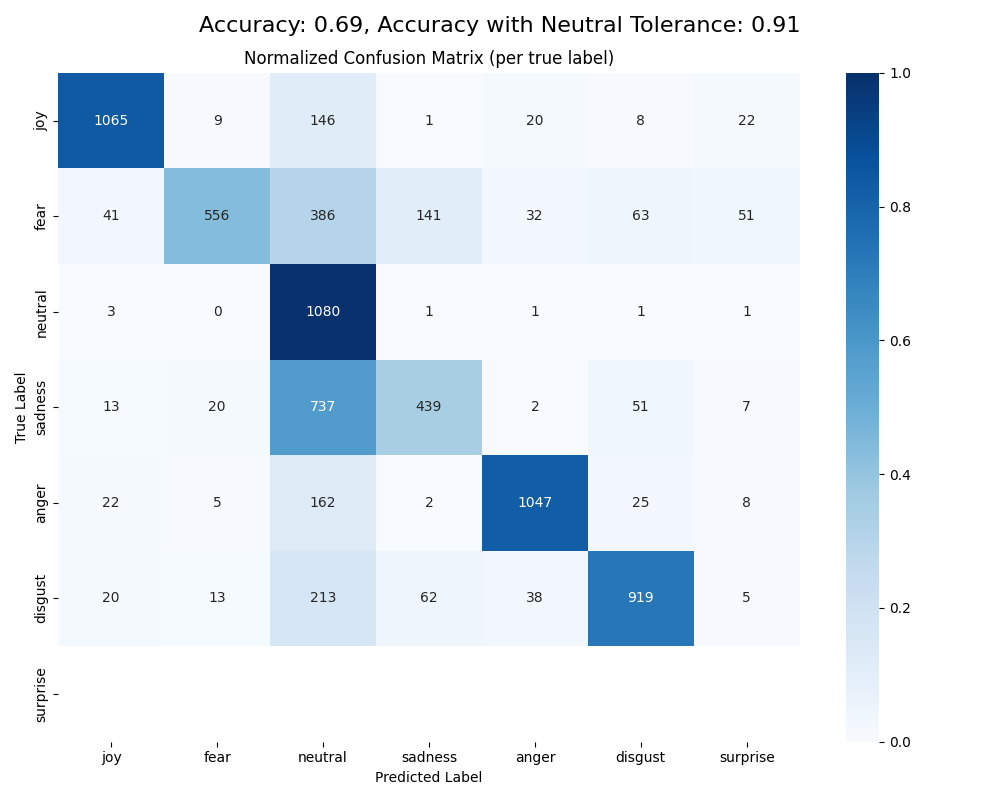}
        \caption{Our pipeline}
        \label{fig:our_comparison}
    \end{subfigure}
    \hfill
    \begin{subfigure}[b]{0.49\textwidth}
        \centering
        \includegraphics[width=\linewidth]{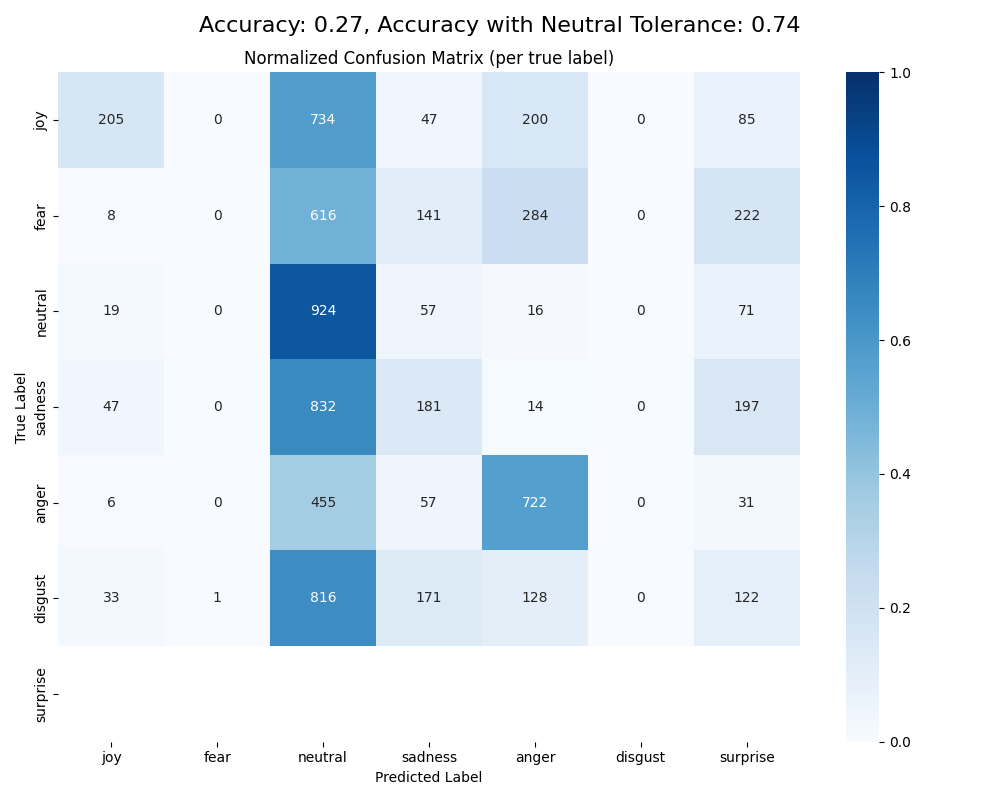}
        \caption{Emotion-Llama}
        \label{fig:emotion_llama_comparision}
    \end{subfigure}
    \caption{Confusion matrices for emotion prediction on CREMA-D.}
    \label{fig:comparision}
\end{figure}

\quad We also compared our results with Emotion-Llama \cite{cheng2024emotion}, an award-winning model for emotion recognition. While Emotion-Llama demonstrates exceptional performance, its large size—approximately six billion parameters, primarily due to its LLaMA backbone—could be considered prohibitively large for edge devices. Ensuring a fair comparison was challenging, as Emotion2Vec \cite{ma2023emotion2vec} and Emotion-Llama were both trained on a diverse range of datasets, raising potential concerns about data contamination. To mitigate this, we selected the CREMA-D dataset for evaluation, since neither of these models was trained on it. As shown in Figure~\ref{fig:comparision}, our pipeline not only surpasses Emotion-Llama on CREMA-D but also remains significantly smaller, with roughly 225 million parameters—about 26 times fewer than Emotion-Llama. In terms of efficiency, inference on our model took only $0.127$ seconds compared to $1.3$ seconds for Emotion-Llama on the same NVIDIA RTX A4000 GPU, making our approach approximately ten times faster.

\section{Discussion}
\quad The experimental results indicate that our multimodal architecture achieves strong generalization across diverse datasets despite relying exclusively on pretrained, off-the-shelf backbones. The integration of Emotion2Vec, DistilRoBERTa, and ResEmotNet allows for consistent performance across speech, text, and visual inputs, while maintaining a total memory footprint compatible with consumer-grade hardware.  

\quad The proposed uncertainty-aware fusion mechanism also proves effective in mitigating conflicts between modalities. By operating directly in the Dirichlet evidence space, our \emph{advanced mitigation} strategy balances modality contributions and reduces overconfident misclassifications. This robustness is particularly valuable in real-world scenarios, where noise, occlusion, or missing data frequently occur. The comparative analysis against the \emph{basic mitigation} method of Wang et al. ~\yrcite{wang2025uefn} demonstrates consistent improvements across both standard and neutral-tolerant accuracy metrics (see Fig.~\ref{fig:accuracy}). Notably, our mitigation framework is modular and model-agnostic: it supports the seamless integration of new input channels (e.g., physiological or contextual signals) and allows individual classifiers to be substituted or retrained independently. Furthermore, since the fusion operates at the level of uncertainty arbitration rather than emotion-specific features, the same principle can be extended beyond emotion recognition to other multimodal reasoning or decision-making tasks. In addition, our pipeline facilitates straightforward parallelization, as each modality can be processed on a separate machine.

\quad Nevertheless, a stronger comparative analysis with state-of-the-art models remains necessary to fully position our approach within the broader landscape of multimodal emotion recognition. While our evaluation against Emotion-Llama offers preliminary insights, it was limited by the availability of fair, non-overlapping datasets. In particular, CREMA-D was selected as an unseen benchmark to avoid data contamination; however, a larger and standardized unseen benchmark would enable more reliable and reproducible cross-model evaluations. Establishing such a benchmark would also help disentangle the effects of architecture design, fusion strategy, and pretraining data coverage. This will also open an avenue to explore MER without fusion, where the individual modalities are treated as features and concatenated to form a high-dimensional emotion descriptor. Leveraging a large, multimodal, dataset of emotions can allow for highly nuanced emotion labeling and detection, which often requires a constructive interplay of emotional modalities. This increased level of nuance can be used to enhance the use-cases of MER, where, for example, human-AI interaction can be modulated to maximize both empathy and instructiveness based on recognized emotion.

\quad Another important direction is the systematic improvement of each modality classifier. While the current implementation leverages existing pretrained backbones, further fine-tuning on emotion-specific datasets or incorporating task-aware loss functions could enhance sensitivity to subtle affective cues. Exploring model compression, cross-modal distillation, or adaptive weighting strategies could also strengthen the contribution of weaker modalities without compromising efficiency. 

\quad Finally, future work should investigate the inclusion of a broader set of emotions and explore how these recognized emotions can be effectively used in downstream tasks. Moving beyond the traditional discrete categories toward more continuous or compound affect representations could capture richer emotional nuances. Moreover, studying how emotion recognition informs decision-making—such as adaptive interfaces, personalized feedback, or mental health monitoring—will be key to ensuring that emotion-aware systems are both impactful and ethically grounded.  

\section{Conclusion}
\quad We introduced a unified multimodal emotion recognition framework that combines efficiency and robustness within a privacy-preserving architecture suitable for real-world deployment. The core contribution of this work lies in a probabilistic fusion mechanism grounded in Dempster–Shafer theory and Dirichlet evidence, providing a principled approach to uncertainty modeling across modalities.

\quad Unlike conventional fusion strategies, our uncertainty-aware mitigation is both \emph{model-agnostic} and \emph{task-agnostic}. It can integrate an arbitrary number of modalities, regardless of their underlying architectures, and can be directly applied to other multimodal reasoning or decision-making tasks beyond emotion recognition. This flexibility makes it a general framework for combining heterogeneous models in any setting where uncertainty and conflict resolution are critical.

\quad Experimental results on multiple benchmark datasets confirm that our system achieves strong performance while maintaining a compact computational footprint, requiring neither extensive retraining nor large-scale GPU resources. Its generalization capabilities, coupled with explicit uncertainty modeling, make it suitable for deployment on consumer-grade devices.

\quad Future work will extend this research along four directions: (1) incorporating temporal modeling for continuous emotion tracking; (2) expanding to new modalities such as physiological or contextual data with potential fusion-less interpretation;  (3) exploring personalized calibration techniques that adapt emotional interpretation to individual users; and (4) integration with speech-to-speech models to add emotional recognition as a feature within the model itself. These developments will further advance the vision of flexible, trustworthy, and empathetic on-device AI systems capable of nuanced human understanding.

\bibliography{example_paper}

@article{ma2023emotion2vec,
  title={emotion2vec: Self-supervised pre-training for speech emotion representation},
  author={Ma, Ziyang and Zheng, Zhisheng and Ye, Jiaxin and Li, Jinchao and Gao, Zhifu and Zhang, Shiliang and Chen, Xie},
  journal={arXiv preprint arXiv:2312.15185},
  year={2023}
}

@inproceedings{he2016deep,
  title={Deep residual learning for image recognition},
  author={He, Kaiming and Zhang, Xiangyu and Ren, Shaoqing and Sun, Jian},
  booktitle={Proceedings of the IEEE conference on computer vision and pattern recognition},
  pages={770--778},
  year={2016}
}

@article{li2019reliable,
  title={Reliable Crowdsourcing and Deep Locality-Preserving Learning for Unconstrained Facial Expression Recognition},
  author={Li, Shan and Deng, Weihong},
  journal={IEEE Transactions on Image Processing},
  volume={28},
  number={1},
  pages={356--370},
  year={2019},
  publisher={IEEE}
}

@article{wang2025uefn,
  title={UEFN: Efficient uncertainty estimation fusion network for reliable multimodal sentiment analysis},
  author={Wang, Shuai and Ratnavelu, K and Bin Shibghatullah, Abdul Samad},
  journal={Applied Intelligence},
  volume={55},
  number={2},
  pages={171},
  year={2025},
  publisher={Springer}
}

@article{Dempster_Shafer_theory,
  title={The Dempster-Shafer theory of evidence},
  author={Gordon, Jean and Shortliffe, Edward H},
  journal={Rule-Based Expert Systems: The MYCIN Experiments of the Stanford Heuristic Programming Project},
  volume={3},
  number={832-838},
  pages={3--4},
  year={1984},
  publisher={Addison-Wesley Reading, Massachusetts}
}

@article{abdullah2021multimodal,
  title={Multimodal emotion recognition using deep learning},
  author={Abdullah, Sharmeen M Saleem Abdullah and Ameen, Siddeeq Y Ameen and Sadeeq, Mohammed AM and Zeebaree, Subhi},
  journal={Journal of Applied Science and Technology Trends},
  volume={2},
  number={01},
  pages={73--79},
  year={2021}
}

@article{adel2023mm,
  title={MM-EMOR: multi-modal emotion recognition of social media using concatenated deep learning networks},
  author={Adel, Omar and Fathalla, Karma M and Abo ElFarag, Ahmed},
  journal={Big Data and Cognitive Computing},
  volume={7},
  number={4},
  pages={164},
  year={2023},
  publisher={MDPI}
}

@inproceedings{xie2013multimodal,
  title={Multimodal information fusion of audiovisual emotion recognition using novel information theoretic tools},
  author={Xie, Zhibing and Guan, Ling},
  booktitle={2013 IEEE International Conference on Multimedia and Expo (ICME)},
  pages={1--6},
  year={2013},
  organization={IEEE}
}

@article{alswaidan2020survey,
  title={A survey of state-of-the-art approaches for emotion recognition in text},
  author={Alswaidan, Nourah and Menai, Mohamed El Bachir},
  journal={Knowledge and Information Systems},
  volume={62},
  number={8},
  pages={2937--2987},
  year={2020},
  publisher={Springer}
}

@inproceedings{devault2014simsensei,
  title={SimSensei Kiosk: A virtual human interviewer for healthcare decision support},
  author={DeVault, David and Artstein, Ron and Benn, Grace and Dey, Teresa and Fast, Ed and Gainer, Alesia and Georgila, Kallirroi and Gratch, Jon and Hartholt, Arno and Lhommet, Margaux and others},
  booktitle={Proceedings of the 2014 international conference on Autonomous agents and multi-agent systems},
  pages={1061--1068},
  year={2014}
}

@article{zhao2021emotion,
  title={Emotion recognition from multiple modalities: Fundamentals and methodologies},
  author={Zhao, Sicheng and Jia, Guoli and Yang, Jufeng and Ding, Guiguang and Keutzer, Kurt},
  journal={IEEE Signal Processing Magazine},
  volume={38},
  number={6},
  pages={59--73},
  year={2021},
  publisher={IEEE}
}

@article{pal2021development,
  title={Development and progress in sensors and technologies for human emotion recognition},
  author={Pal, Shantanu and Mukhopadhyay, Subhas and Suryadevara, Nagender},
  journal={Sensors},
  volume={21},
  number={16},
  pages={5554},
  year={2021},
  publisher={MDPI}
}

@article{russell1980circumplex,
  title={A circumplex model of affect.},
  author={Russell, James A},
  journal={Journal of personality and social psychology},
  volume={39},
  number={6},
  pages={1161},
  year={1980},
  publisher={American Psychological Association}
}

@article{mishra2024speech,
  title={Speech emotion recognition using mfcc-based entropy feature},
  author={Mishra, Siba Prasad and Warule, Pankaj and Deb, Suman},
  journal={Signal, image and video processing},
  volume={18},
  number={1},
  pages={153--161},
  year={2024},
  publisher={Springer}
}

@article{koolagudi2012emotion,
  title={Emotion recognition from speech using sub-syllabic and pitch synchronous spectral features},
  author={Koolagudi, Shashidhar G and Krothapalli, Sreenivasa Rao},
  journal={International Journal of Speech Technology},
  volume={15},
  number={4},
  pages={495--511},
  year={2012},
  publisher={Springer}
}

@inproceedings{koolagudi2009spectral,
  title={Spectral features for emotion classification},
  author={Koolagudi, Shashidhar G and Nandy, Sourav and Rao, K Sreenivasa},
  booktitle={2009 IEEE International Advance Computing Conference},
  pages={1292--1296},
  year={2009},
  organization={IEEE}
}

@inproceedings{huang2014speech,
  title={Speech emotion recognition using CNN},
  author={Huang, Zhengwei and Dong, Ming and Mao, Qirong and Zhan, Yongzhao},
  booktitle={Proceedings of the 22nd ACM international conference on Multimedia},
  pages={801--804},
  year={2014}
}

@article{zhang2019spontaneous,
  title={Spontaneous speech emotion recognition using multiscale deep convolutional LSTM},
  author={Zhang, Shiqing and Zhao, Xiaoming and Tian, Qi},
  journal={IEEE Transactions on Affective Computing},
  volume={13},
  number={2},
  pages={680--688},
  year={2019},
  publisher={IEEE}
}

@article{lieskovska2021review,
  title={A review on speech emotion recognition using deep learning and attention mechanism},
  author={Lieskovsk{\'a}, Eva and Jakubec, Maro{\v{s}} and Jarina, Roman and Chmul{\'\i}k, Michal},
  journal={Electronics},
  volume={10},
  number={10},
  pages={1163},
  year={2021},
  publisher={MDPI}
}

@inproceedings{tarantino2019self,
  title={Self-attention for speech emotion recognition.},
  author={Tarantino, Lorenzo and Garner, Philip N and Lazaridis, Alexandros and others},
  booktitle={Interspeech},
  pages={2578--2582},
  year={2019}
}

@article{roy2024resemotenet,
  title={ResEmoteNet: bridging accuracy and loss reduction in facial emotion recognition},
  author={Roy, Arnab Kumar and Kathania, Hemant Kumar and Sharma, Adhitiya and Dey, Abhishek and Ansari, Md Sarfaraj Alam},
  journal={IEEE Signal Processing Letters},
  year={2024},
  publisher={IEEE}
}

@article{ning2024representation,
  title={Representation learning and identity adversarial training for facial behavior understanding},
  author={Ning, Mang and Salah, Albert Ali and Ertugrul, Itir Onal},
  journal={arXiv preprint arXiv:2407.11243},
  year={2024}
}

@inproceedings{zhao2021robust,
  title={Robust lightweight facial expression recognition network with label distribution training},
  author={Zhao, Zengqun and Liu, Qingshan and Zhou, Feng},
  booktitle={Proceedings of the AAAI conference on artificial intelligence},
  volume={35},
  number={4},
  pages={3510--3519},
  year={2021}
}

@article{zhang2023dual,
  title={A dual-direction attention mixed feature network for facial expression recognition},
  author={Zhang, Saining and Zhang, Yuhang and Zhang, Ye and Wang, Yufei and Song, Zhigang},
  journal={Electronics},
  volume={12},
  number={17},
  pages={3595},
  year={2023},
  publisher={MDPI}
}

@misc{githubGitHubSithu31296EasyFace,
	author = {Sithu Aung},
	title = {{G}it{H}ub - sithu31296/{E}asy{F}ace: {E}asy-to-use {F}ace {A}nalysis {T}ool --- github.com},
	howpublished = {\url{https://github.com/sithu31296/EasyFace/tree/main?tab=readme-ov-file#references}},
	year = {2022},
	note = {[Accessed 26-09-2025]},
}

@article{liu2019roberta,
  title={Roberta: A robustly optimized bert pretraining approach},
  author={Liu, Yinhan and Ott, Myle and Goyal, Naman and Du, Jingfei and Joshi, Mandar and Chen, Danqi and Levy, Omer and Lewis, Mike and Zettlemoyer, Luke and Stoyanov, Veselin},
  journal={arXiv preprint arXiv:1907.11692},
  year={2019}
}

@article{cortiz2021exploring,
  title={Exploring transformers in emotion recognition: a comparison of bert, distillbert, roberta, xlnet and electra},
  author={Cortiz, Diogo},
  journal={arXiv preprint arXiv:2104.02041},
  year={2021}
}

@inproceedings{adoma2020comparative,
  title={Comparative analyses of bert, roberta, distilbert, and xlnet for text-based emotion recognition},
  author={Adoma, Acheampong Francisca and Henry, Nunoo-Mensah and Chen, Wenyu},
  booktitle={2020 17th international computer conference on wavelet active media technology and information processing (ICCWAMTIP)},
  pages={117--121},
  year={2020},
  organization={IEEE}
}

@article{barcena2025textual,
  title={Textual emotion detection with complementary BERT transformers in a Condorcet’s Jury theorem assembly},
  author={B{\'a}rcena Ruiz, Gerardo and Gil Herrera, Richard De Jes{\'u}s},
  journal={Knowledge-Based Systems},
  volume={326},
  year={2025},
  publisher={Elsevier}
}

@article{alqarni2025emotion,
  title={Emotion-Aware RoBERTa enhanced with emotion-specific attention and TF-IDF gating for fine-grained emotion recognition},
  author={Alqarni, Fatimah and Sagheer, Alaa and Alabbad, Amira and Hamdoun, Hala},
  journal={Scientific Reports},
  volume={15},
  number={1},
  pages={17617},
  year={2025},
  publisher={Nature Publishing Group UK London}
}

@article{khan2025memocmt,
  title={MemoCMT: multimodal emotion recognition using cross-modal transformer-based feature fusion},
  author={Khan, Mustaqeem and Tran, Phuong-Nam and Pham, Nhat Truong and El Saddik, Abdulmotaleb and Othmani, Alice},
  journal={Scientific reports},
  volume={15},
  number={1},
  pages={5473},
  year={2025},
  publisher={Nature Publishing Group UK London}
}

@article{alqurashi2024decision,
  title={Decision fusion based multimodal hierarchical method for speech emotion recognition from audio and text},
  author={Alqurashi, Nawal and Li, Yuhua and Sidorov, Kirill and Marshall, Andrew},
  journal={European Symposium on Artificial Neural Networks, Computational Intelligence and
Machine Learning},
  year={2024}
}

@article{karani2022review,
  title={Review on multimodal fusion techniques for human emotion recognition},
  author={Karani, Ruhina and Desai, Sharmishta},
  journal={Int. J. Adv. Comput. Sci. Appl},
  volume={13},
  pages={287--296},
  year={2022}
}

@article{xu2020emotion,
  title={Emotion recognition model based on the Dempster--Shafer evidence theory},
  author={Xu, Qihua and Zhang, Chunyue and Sun, Bo},
  journal={Journal of Electronic Imaging},
  volume={29},
  number={2},
  pages={023018--023018},
  year={2020},
  publisher={Society of Photo-Optical Instrumentation Engineers}
}

@inproceedings{martin2006enterface,
  title={The eNTERFACE'05 audio-visual emotion database},
  author={Martin, Olivier and Kotsia, Irene and Macq, Benoit and Pitas, Ioannis},
  booktitle={22nd international conference on data engineering workshops (ICDEW'06)},
  pages={8--8},
  year={2006},
  organization={IEEE}
}

@inproceedings{wang2020mead,
  title={Mead: A large-scale audio-visual dataset for emotional talking-face generation},
  author={Wang, Kaisiyuan and Wu, Qianyi and Song, Linsen and Yang, Zhuoqian and Wu, Wayne and Qian, Chen and He, Ran and Qiao, Yu and Loy, Chen Change},
  booktitle={European conference on computer vision},
  pages={700--717},
  year={2020},
  organization={Springer}
}

@article{poria2018meld,
  title={Meld: A multimodal multi-party dataset for emotion recognition in conversations},
  author={Poria, Soujanya and Hazarika, Devamanyu and Majumder, Navonil and Naik, Gautam and Cambria, Erik and Mihalcea, Rada},
  journal={arXiv preprint arXiv:1810.02508},
  year={2018}
}

@article{livingstone2018ryerson,
  title={The Ryerson Audio-Visual Database of Emotional Speech and Song (RAVDESS): A dynamic, multimodal set of facial and vocal expressions in North American English},
  author={Livingstone, Steven R and Russo, Frank A},
  journal={PloS one},
  volume={13},
  number={5},
  pages={e0196391},
  year={2018},
  publisher={Public Library of Science San Francisco, CA USA}
}

@article{cao2014crema,
  title={Crema-d: Crowd-sourced emotional multimodal actors dataset},
  author={Cao, Houwei and Cooper, David G and Keutmann, Michael K and Gur, Ruben C and Nenkova, Ani and Verma, Ragini},
  journal={IEEE transactions on affective computing},
  volume={5},
  number={4},
  pages={377--390},
  year={2014},
  publisher={IEEE}
}

@inproceedings{deng2020retinaface,
  title={Retinaface: Single-shot multi-level face localisation in the wild},
  author={Deng, Jiankang and Guo, Jia and Ververas, Evangelos and Kotsia, Irene and Zafeiriou, Stefanos},
  booktitle={Proceedings of the IEEE/CVF conference on computer vision and pattern recognition},
  pages={5203--5212},
  year={2020}
}

@article{ozdemir2017copula,
  title={Copula based classifier fusion under statistical dependence},
  author={Ozdemir, Onur and Allen, Thomas G and Choi, Sora and Wimalajeewa, Thakshila and Varshney, Pramod K},
  journal={IEEE Transactions on Pattern Analysis and Machine Intelligence},
  volume={40},
  number={11},
  pages={2740--2748},
  year={2017},
  publisher={IEEE}
}

@article{cheng2024emotion,
  title={Emotion-llama: Multimodal emotion recognition and reasoning with instruction tuning},
  author={Cheng, Zebang and Cheng, Zhi-Qi and He, Jun-Yan and Wang, Kai and Lin, Yuxiang and Lian, Zheng and Peng, Xiaojiang and Hauptmann, Alexander},
  journal={Advances in Neural Information Processing Systems},
  volume={37},
  pages={110805--110853},
  year={2024}
}

@article{ekman1992argument,
  title={An argument for basic emotions},
  author={Ekman, Paul},
  journal={Cognition \& Emotion},
  volume={6},
  number={3-4},
  pages={169--200},
  year={1992},
  publisher={Taylor \& Francis}
}

@inproceedings{sanh2019distilbert,
  title={DistilBERT, a distilled version of BERT: smaller, faster, cheaper and lighter},
  author={Sanh, Victor and Debut, Lysandre and Chaumond, Julien and Wolf, Thomas},
  booktitle={NeurIPS EMC2 Workshop},
  year={2019}
}

@inproceedings{demszky2020goemotions,
  title={GoEmotions: A Dataset of Fine-Grained Emotions},
  author={Demszky, Dorottya and others},
  booktitle={ACL},
  year={2020}
}

@inproceedings{micikevicius2018mixed,
  title={Mixed precision training},
  author={Micikevicius, Paulius and others},
  booktitle={ICLR},
  year={2018}
}

@article{imagenet15russakovsky,
    Author = {Olga Russakovsky and Jia Deng and Hao Su and Jonathan Krause and Sanjeev Satheesh and Sean Ma and Zhiheng Huang and Andrej Karpathy and Aditya Khosla and Michael Bernstein and Alexander C. Berg and Li Fei-Fei},
    Title = { {ImageNet Large Scale Visual Recognition Challenge} },
    Year = {2015},
    journal   = {International Journal of Computer Vision (IJCV)},
    doi = {10.1007/s11263-015-0816-y},
    volume={115},
    number={3},
    pages={211-252}
}

@BOOK{Lewin1936,
  title    = "Principles of Topological Psychology",
  author   = "Lewin, Kurt",
  publisher = "McGraw-Hill Book Company, Inc",
  year     =  1936,
  language = "en",
  page = 12
}
\bibliographystyle{mlsys2025}



\end{document}